\documentclass[10pt,twocolumn,letterpaper]{article}

\usepackage[pagenumbers]{cvpr} 
\usepackage[accsupp]{axessibility}  

\usepackage{amsmath}
\usepackage{amssymb}

\usepackage[pagebackref,breaklinks,colorlinks]{hyperref}

\usepackage[capitalize]{cleveref}
\crefname{section}{Sec.}{Secs.}
\Crefname{section}{Section}{Sections}
\Crefname{table}{Table}{Tables}
\crefname{table}{Tab.}{Tabs.}

\usepackage[utf8]{inputenc} 
\usepackage[T1]{fontenc}    
\usepackage{hyperref}       
\usepackage{url}            
\usepackage{booktabs}       
\usepackage{amsfonts}       
\usepackage{nicefrac}       
\usepackage{microtype}      
\usepackage{xcolor}         
\usepackage{array}
\usepackage{graphicx}
\usepackage{mathrsfs}
\usepackage{pifont}
\usepackage{comment}
\usepackage{tabulary,multirow,xspace}
\usepackage{fixmath,mathtools,mmstyle}
\usepackage{cite}
\usepackage{subcaption}
\usepackage{caption}
\usepackage{float}
\newcommand{\myparagraph}[1]{{\noindent\bf #1}}

\def\cvprPaperID{6268} 
\def\confName{CVPR}
\def\confYear{2022}

\begin{document}

\title{CRIS: CLIP-Driven Referring Image Segmentation}

\author{
  Zhaoqing Wang$^{1,2}\thanks{Equal technical contribution.}$\quad
  Yu Lu$^{3\ast}$\quad
  Qiang Li$^{4\ast}$\quad
  Xunqiang Tao$^3$\quad
  Yandong Guo$^3$\\
  Mingming Gong$^5$\quad
  Tongliang Liu$^1$\\
  \small{$^1$University of Sydney};
  \small{$^2$OPPO Research Institute};
  \small{$^3$Beijing University of Posts and Telecommunications}\\
  \small{$^4$Kuaishou Technology};
  \small{$^5$University of Melbourne}\\
  \tt\small \{derrickwang005,leetsiang.cloud,taoxunqiang\}@gmail.com;
  \tt\small{aniki@bupt.edu.cn}\\
  \tt\small{yandong.guo@live.com};
  \tt\small{mingming.gong@unimelb.edu.au};
  \tt\small{tongliang.liu@sydney.edu.au}
}
\maketitle

\begin{abstract}
    Referring image segmentation aims to segment a referent via a natural linguistic expression.
    Due to the distinct data properties between text and image, it is challenging for a network to well align text and pixel-level features.
    Existing approaches use pretrained models to facilitate learning, yet separately transfer the language/vision knowledge from pretrained models, ignoring the multi-modal corresponding information.
    Inspired by the recent advance in Contrastive Language-Image Pretraining (CLIP), in this paper, we propose an end-to-end CLIP-Driven Referring Image Segmentation framework (CRIS).
    To transfer the multi-modal knowledge effectively, CRIS resorts to vision-language decoding and contrastive learning for achieving the text-to-pixel alignment.
    More specifically, we design a vision-language decoder to propagate fine-grained semantic information from textual representations to each pixel-level activation, which promotes consistency between the two modalities.
    In addition, we present text-to-pixel contrastive learning to explicitly enforce the text feature similar to the related pixel-level features and dissimilar to the irrelevances.
    The experimental results on three benchmark datasets demonstrate that our proposed framework significantly outperforms the state-of-the-art performance without any post-processing.
    The code will be released.
\end{abstract}

\begin{figure}[t]
	\centering
    \includegraphics[width=0.88\linewidth]{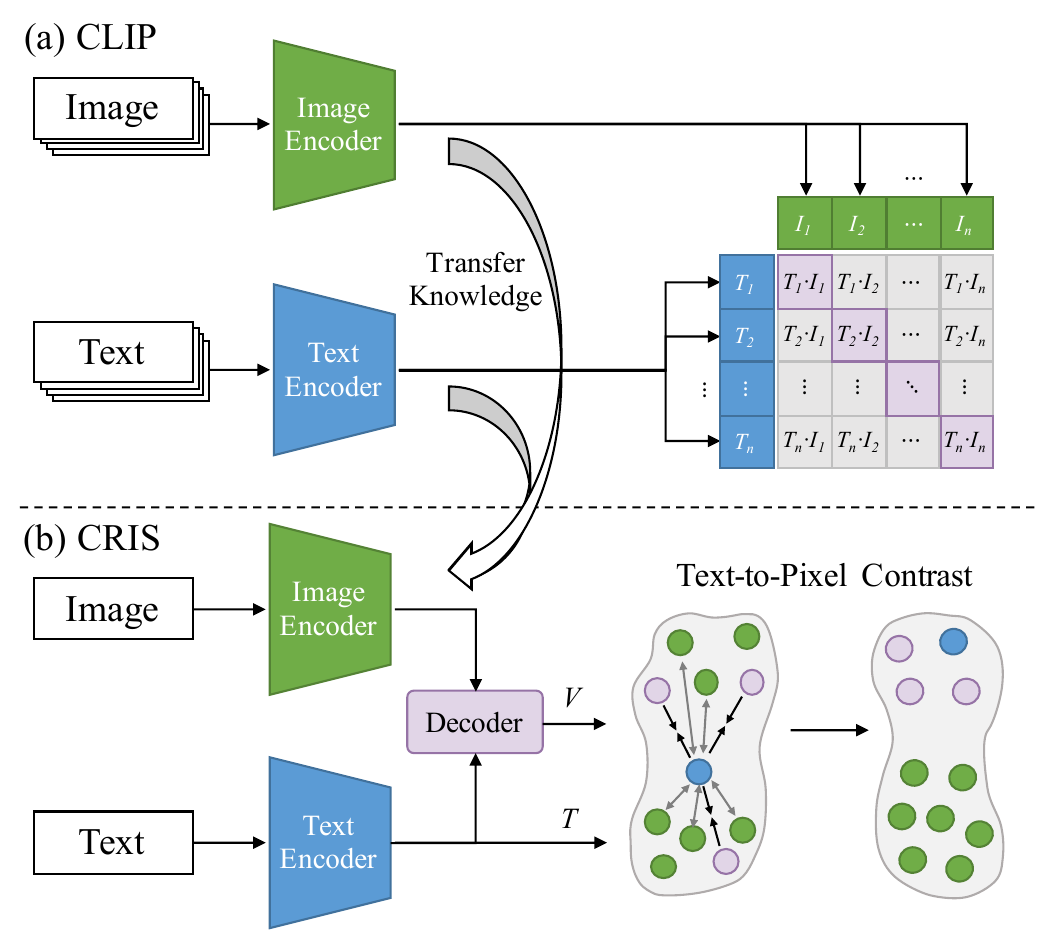}
	\caption{\textbf{An illustration of our main idea.}
            (a) CLIP \cite{radford2021learning} jointly trains an image encoder and a text encoder to predict the correct pairings of a batch of image \textcolor{green}{$I$} and text \textcolor{blue}{$T$}, which can capture the multi-modal corresponding information.
            (b) To transfer this knowledge of the CLIP model from image level to pixel level, we propose a CLIP-Driven Referring Image Segmentation (CRIS) framework.
            Firstly, we design a \textcolor[RGB]{159,96,187}{vision-language decoder} to propagate fine-grained semantic information from textual features to pixel-level visual features.
            Secondly, we combine all pixel-level visual features \textcolor{green}{$V$} with the global textual feature \textcolor{blue}{$T$} and adopt contrastive learning to pull text and \textcolor[RGB]{159,96,187}{related pixel-wise features} closer and push \textcolor{green}{other irrelevances} away.}
            \label{fig:Introduction}
	\vspace{-0.2cm}
\end{figure}

\section{Introduction}\label{sec:Introduction}
Referring image segmentation \cite{hu2016segmentation,yu2018mattnet,ye2019cross} is a fundamental and challenging task at the intersection of vision and language understanding,
which can be potentially used in a wide range of applications, including interactive image editing and human-object interaction.
Unlike semantic and instance segmentation \cite{fu2019dual,he2019adaptive,wu2020ginet,he2017mask}, which requires segmenting the visual entities belonging to a pre-determined set of categories,
referring image segmentation is not limited to indicating specific categories but finding a particular region according to the input language expression.

\begin{figure*}[t]
    \begin{center}
        \includegraphics[width=0.95\linewidth]{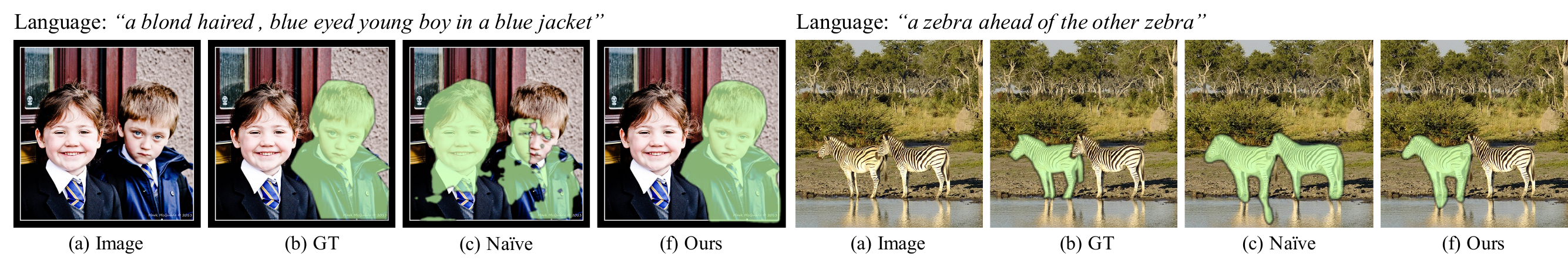}
        \caption{\textbf{Comparison between the direct fine-tuning and our proposed methods.}
                 ``Naive'' denotes the direct fine-tuning mentioned in section \ref{sec:Experiments}.
                 Compared with the direct fine-tuning,
                 our method can not only leverage the powerful cross-modal matching capability of the CLIP,
                 but also learn fine-grained visual representations.}
    \label{fig:compare}
    \end{center}
    \vspace{-7.0mm}
\end{figure*}

Since the image and language modality maintain different properties, it is difficult to explicitly align textual features with pixel-level activations.
Benefiting from the powerful capacity of the deep neural network, early approaches \cite{hu2016segmentation,liu2017recurrent,li2018referring,margffoy2018dynamic} concatenate textual features with each visual activation directly, and use these combined features to generate the segmentation mask.
Subsequently, to address the lack of adequate interaction between two modalities, a series of methods \cite{chen2019referring,shi2018key,ye2019cross,huang2020referring,huilinguistic} adopt the language-vision attention mechanism to better learn cross-modal features.

Existing methods \cite{chen2019referring,shi2018key,ye2019cross,huang2020referring,huilinguistic} leverage external knowledge to facilitate learning in common, while they mainly utilize a single-modal pretraining (\eg, the pretrained image or text encoder), which is short of multi-modal correspondence information.
By resorting to language supervision from large-scale unlabeled data, vision-language pretraining \cite{radford2021learning,wang2021simvlm,miech2020end} is able to learn ample multi-modal representations.
Recently, the remarkable success of the CLIP \cite{radford2021learning} has shown its capability of learning SOTA image-level visual concepts from 400 million image-text pairs,
which assists many multi-modal tasks achieve significant improvements, including image-text retrieval \cite{radford2021learning}, video-text retrieval \cite{luo2021clip4clip,fang2021clip2video}.
However, as shown in Figure~\ref{fig:compare}, the direct usage of the CLIP can be sub-optimal for pixel-level prediction tasks, \eg, referring image segmentation, duo to the discrepancy between image-level and pixel-level prediction.
The former focuses on the global information of an input image, while the latter needs to learn fine-grained visual representations for each spatial activation.

In this paper, we explore leveraging the powerful knowledge of the CLIP model for referring image segmentation, in order to enhance the ability of cross-modal matching.
Considering the characteristics of referring image segmentation, we propose an effective and flexible framework named CLIP-Driven Referring Image Segmentation (CRIS), which can transfer ample multi-modal corresponding knowledge of the CLIP for achieving text-to-pixel alignment.
Firstly, we propose a visual-language decoder that captures long-range dependencies of pixel-level features through the self-attention operation and adaptively propagate fine-structured textual features into pixel-level features through the cross-attention operation.
Secondly, we introduce the text-to-pixel contrastive learning,
which can align linguistic features and the corresponding pixel-level features, meanwhile distinguishing irrelevant pixel-level features in the multi-modal embedding space.
Based on this scheme, the model can explicitly learn fine-grained visual concepts by interwinding the linguistic and pixel-level visual features.


Our main contributions are summarized as follow:
\begin{itemize}
    \item We propose a CLIP-Driven Referring Image Segmentation framework (CRIS) to transfer the knowledge of the CLIP model for achieving text-to-pixel alignment.
    \item We take fully advantage of this multi-modal knowledge with two innovative designs, \ie, the vision-language decoder and text-to-pixel contrastive learning.
    \item The experimental results on three challenging benchmarks significantly outperform previous state-of-the-art methods by large margins (\eg, $+\,4.89\,IoU$ on RefCOCO, $+\,8.88\,IoU$ on RefCOCO+, $+\,5.47\,IoU$ on G-Ref).
\end{itemize}
\section{Related Work}

\myparagraph{Vision-Language Pretraining.}
Vision-Language pretraining has made rapid progress in recent years and achieved impressive performance on various multi-modal downstream tasks.
By resorting to semantic supervision from large-scale image data, several approaches \cite{radford2021learning,wang2021simvlm,miech2020end} were proposed to learn visual representations from text representations. MIL-NCE \cite{miech2020end} mainly explored leveraging noisy large-scale Howto100M \cite{miech2019howto100m} instructional videos to learn a better video encoder via an end-to-end manner.
SimVLM \cite{wang2021simvlm} reduced the training complexity by leveraging large-scale weak supervision, and adopted a single prefix language modeling objective in an end-to-end manner.
Benefit from the large-scale image and text pairs collected from the Internet, a recent approach, \ie, Contrastive Language-Image Pretraining (CLIP) \cite{radford2021learning}, achieved the notable success of aligning two modalities representations in the embedding space.
CLIP adopted contrastive learning with high-capacity language models and visual feature encoders to capture compelling visual concepts for zero-shot image classification.
More recently, a series of works  \cite{tang2021clip4caption,luo2021clip4clip,fang2021clip2video,patashnik2021styleclip} were proposed to transfer the knowledge of CLIP models to downstream tasks and achieved promising results, such as video caption, video-text retrieval, and image generation.
Different from these works, we transfer these image-level visual concepts to referring image segmentation for leveraging multi-modal corresponding information.

\myparagraph{Contrastive Learning}
Date back to \cite{hadsell2006dimensionality}, these approaches learned representations by contrasting positive pairs against negative pairs.
Several approaches \cite{he2020momentum,chen2020improved,chen2020simple,wu2018unsupervised,li2022selective} were proposed to treat each image as a class and use contrastive loss-based instance discrimination for representation learning.
Recently, VADeR and DenseCL \cite{wang2020DenseCL,pinheiro2020unsupervised} proposed to explore pixel-level contrastive learning to fill the gap between self-supervised representation learning and dense prediction tasks.
Besides, CLIP \cite{radford2021learning} proposed a promising alternative that directly learns transferable visual concepts from large-scale collected image-text pairs by using cross-modal contrastive loss.
In this paper, we propose a CLIP-Driven Referring Image Segmentation (CRIS) framework to transfer the knowledge of the CLIP model to referring image segmentation in an end-to-end manner.

\begin{figure*}[htbp]
    \begin{center}
        \includegraphics[width=0.9\linewidth]{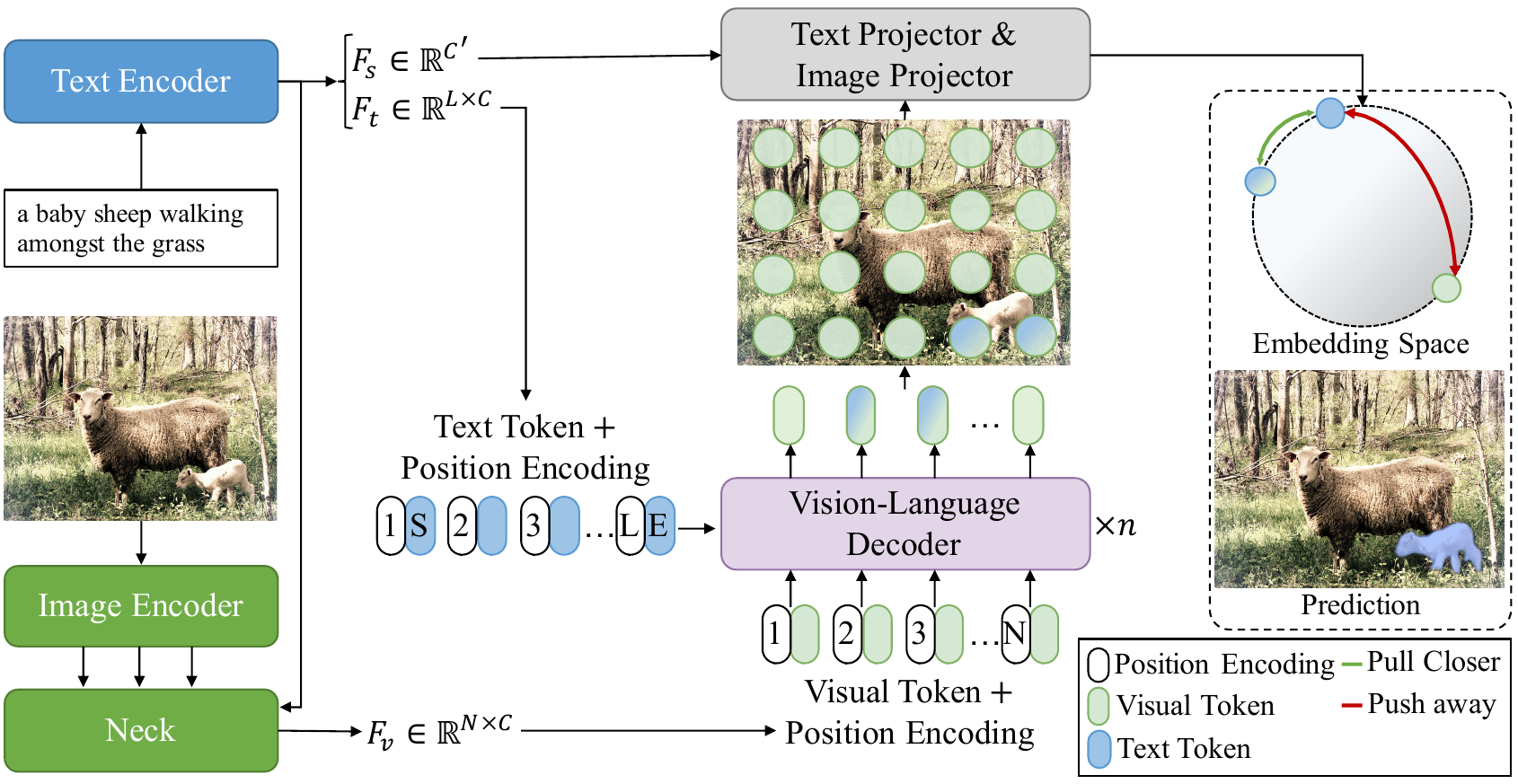}
        \caption{\textbf{The overview of the proposed CLIP-Driven Referring Image Segmentation (CRIS) framework.}
                 CRIS mainly consists of a text encoder, an image encoder, a cross-modal neck, a vision-language decoder, and two projectors.
                 The vision-language decoder is used to adaptively propagate semantic information from textual features to visual features.
                 The text-to-pixel contrastive learning is used to explicitly learn fine-grained multi-modal corresponding information by interwinding the text features and pixel-level visual features.}
    \label{fig:method}
    \end{center}
    \vspace{-7mm}
\end{figure*}

\myparagraph{Referring Image Segmentation}
Referring image segmentation is to segment a target region (\eg, object or stuff) in an image by understanding a given natural linguistic expression, which was first introduced by \cite{hu2016segmentation}.
Early works \cite{liu2017recurrent,li2018referring,margffoy2018dynamic} first extracted visual and linguistic features by CNN and LSTM, respectively, and directly concatenated two modalities to obtain final segmentation results by a FCN \cite{long2015fully}.
In \cite{yu2018mattnet}, they proposed a two-stage method that first extracted instances using Mask R-CNN \cite{he2017mask}, and then adopted linguistic features to choose the target from those instances.
Besides, MCN \cite{luo2020multi} designed a framework achieving impressive results.
They learned to optimize two related tasks, \ie, referring expression comprehension and segmentation, simultaneously.

As the attention mechanism arouses more and more interests, a series of works are proposed to adopt the attention mechanism.
It is powerful to extract the visual contents corresponding to the language expression.
\cite{shi2018key} used the vision-guided linguistic attention to aggregate the linguistic context of each visual region adaptively.
\cite{ye2019cross} designed a Cross-Modal Self-Attention (CSMA) module to focus on informative words in the sentence and crucial regions in the image.
\cite{hu2020bi} proposed a bi-directional relationship inferring network that adopted a language-guided visual and vision-guided linguistic attention module to capture the mutual guidance between two modalities.
Besides, LTS \cite{jing2021locate} designs a strong pipeline that decouples the task into a ``Locate-Then-Segment'' scheme by introducing the position prior.
EFNet \cite{feng2021encoder} designs a co-attention mechanism to use language to refine the multi-modal features progressively,
which can promote the consistent of the cross-modal information representation.
More recently, VLT\cite{ding2021vlt} employs transformer to build a network with an encoder-decoder attention mechanism for enhancing the global context information.
Different from previous methods, we aim to leverage the knowledge of the CLIP, in order to improving the compatibility of multi-modal information and boost the ability of cross-modal matching.


\section{Methodology}\label{sec:methods}
As illustrated in Figure~\ref{fig:method}, we introduce how the proposed CRIS framework transfers the knowledge of CLIP to referring image segmentation to achieve text-to-pixel alignment by leveraging multi-modal corresponding information.
Firstly, we use a ResNet \cite{he2016deep} and a Transformer \cite{vaswani2017attention} to extract image and text features respectively, which are further fused to obtain the simple multi-modal features.
Secondly, these features and text features are fed into the vision-language decoder to propagate fine-grained semantic information from textual representations to pixel-level visual activations.
Finally, we use two projectors to produce the final prediction mask, and adopt the text-to-pixel contrastive loss to explicitly align the text features with the relevant pixel-level visual features.

\subsection{Image \& Text Feature Extraction}
As illustrated in Figure~\ref{fig:method}, the input of our framework consists of an image $I$ and a referring expression $T$.

\textbf{Image Encoder.} 
For an input image $I \in \mathbb{R}^{H \times W \times 3}$, we utilize multiple visual features from the 2th-4th stages of the ResNet,
which are defined as $F_{v2} \in \mathbb{R}^{\frac{H}{8} \times \frac{W}{8} \times C_2}$, $F_{v3} \in \mathbb{R}^{\frac{H}{16} \times \frac{W}{16} \times C_3}$, and $F_{v4} \in \mathbb{R}^{\frac{H}{32} \times \frac{W}{32} \times C_4}$, respectively.
Note that $C$ is the feature dimension, $H$ and $W$ are the height and width of the original image.

\textbf{Text Encoder.}
For an input expression $T \in \mathbb{R}^{L}$, we adopt a Transformer \cite{vaswani2017attention} modified by \cite{radford2019language} to extract text features $F_t \in \mathbb{R}^{L \times C}$.
The Transformer operates on a lower-cased byte pair encoding (BPE) representation of the text with a 49,152 vocab size \cite{sennrich2015neural},
and the text sequence is bracketed with \texttt{[SOS]} and \texttt{[EOS]} tokens.
The activations of the highest layer of the transformer at the \texttt{[EOS]} token are further transformed as the global textual representation $F_s \in \mathbb{R}^{C^{\prime}}$.
Note that $C$ and $C^{\prime}$ are the feature dimension, $L$ is the length of the referring expression.

\textbf{Cross-modal Neck.}
Given multiple visual features and the global textual representation $F_{s}$, we obtain the simple multi-modal feature $F_{m4} \in \mathbb{R}^{\frac{H}{16} \times \frac{W}{16} \times C}$ by fusing $F_{v4}$ with $F_{s}$:
\begin{equation}
    F_{m4} = Up(\sigma(F_{v4}W_{v4}) \cdot \sigma(F_{s}W_{s})),
    \label{eq1}
\end{equation}
where $Up(\cdot)$ denotes $2\times$ upsampling, $\cdot$ denotes the element-wise multiplication, $\sigma$ denotes ReLU, $W_{v4}$ and $W_{s}$ are two learnable matrices to transform the visual and textual representations into the same feature dimension.
Then, the multi-modal features $F_{m2}$ and $F_{m3}$ are obtained by:
\begin{equation}
\begin{aligned}
    &F_{m_3} = [\sigma(F_{m_4}W_{m_4}),\, \sigma(F_{v_3}W_{v_3})],\\
    &F_{m_2} = [\sigma(F_{m_3}W_{m_3}),\, \sigma(F_{v_2}^{\prime}W_{v_2})],\,F_{v_{2}}^{\prime} = Avg(F_{v_2}),
    \label{eq2}
\end{aligned}
\end{equation}
where $Avg(\cdot)$ denotes a kernel size of $2 \times 2$ average pooling with 2 strides, respectively. $[,]$ is the concatenation operation.
Subsequently, we concatenate three multi-modal features and use a $1 \times 1$ convolution layer to aggregate them:
\begin{equation}
    F_{m} = Conv([F_{m_2},\,F_{m_3},\,F_{m_4}]),
    \label{eq3}
\end{equation}
where $F_{m} \in \mathbb{R}^{\frac{H}{16} \times \frac{W}{16} \times C}$.
Finally, we concatenate a 2D spatial coordinate feature $F_{coord} \in \mathbb{R}^{\frac{H}{16} \times \frac{W}{16} \times 2}$ with $F_{m}$ and fuse that by a $3 \times 3$ convolution \cite{liu2018coordconv}. The visual feature $F_{v} \in \mathbb{R}^{\frac{H}{16} \times \frac{W}{16} \times C}$ is calculated as follow,
\begin{equation}
    F_{v} = Conv([F_{m},\, F_{coord}]).
    \label{eq4}
\end{equation}
As shown in figure~\ref{fig:method}, we flatten the spatial domain of $F_{v}$ into a sequence, forming the visual feature $F_{v} \in \mathbb{R}^{N \times C},\,N=\frac{H}{16} \times \frac{W}{16}$, which is utilized in the following process.

\subsection{Vision-Language Decoder}
We design a vision-language decoder to adaptively propagate fine-grained semantic information from textual features to visual features.
As shown in Figure~\ref{fig:method}, the decoder module takes textual features $F_{t} \in \mathbb{R}^{L \times C}$ and pixel-level visual features $F_{v} \in \mathbb{R}^{N \times C}$ as inputs, which can provide ample textual information corresponding to visual features.
To capture positional information, the fixed sine spatial positional encodings are added to $F_{v}$ \cite{carion2020end} and $F_{t}$ \cite{vaswani2017attention}, respectively.
The vision-language decoder composed of $n$ layers is applied to generate a sequence of evolved multi-modal features $F_c \in \mathbb{R}^{N \times C}$.
Following the standard architecture of the transformer \cite{vaswani2017attention}, each layer consists of a multi-head self-attention layer, a multi-head cross-attention layer, and a feed-forward network.
In one decoder layer, $F_{v}$ is first sent into the multi-head self-attention layer to capture global contextual information:
\begin{equation}
    F_{v}^{\prime} = MHSA(LN(F_{v})) + F_{v},
    \label{eq5}
\end{equation}
where $F_{v}^{\prime}$ is the evolved visual feature, $MHSA(\cdot)$ and $LN(\cdot)$ denote the multi-head self-attention layer and Layer Normalization \cite{ba2016layer}, respectively.
The multi-head self-attention mechanism is composed of three point-wise linear layers mapping $F_{v}$ to intermediate representations, queries $Q \in \mathbb{R}^{N \times d_q}$, keys $K \in \mathbb{R}^{N \times d_k}$ and values $V \in \mathbb{R}^{N \times d_v}$.
Multi-head self-attention is then calculated as follows,
\begin{equation}
    MHSA(Q,\, K,\, V) = softmax(\frac{QK^{T}}{\sqrt{d_k}})V.
    \label{eq6}
\end{equation}
After that, the multi-head cross-attention layer is adopted to propagate fine-grained semantic information into the evolved visual features,
where one point-wise linear layer maps $F_{v}^{\prime}$ to $Q$, and the other two linear layers map $F_{t}$ to $K$ and $V$.
To obtain the multi-modal feature $F_{c}$, the output query $Q$ is further computed by a MLP block of two layers with Layer Normalization and residual connections:
\begin{equation}
\begin{aligned}
    &F_{c}^{\prime} = MHCA(LN(F_{v}^{\prime}),\,F_t) + F_{v}^{\prime},\\
    &F_{c} = MLP(LN(F_{c}^{\prime})) + F_{c}^{\prime},
    \label{eq7}
\end{aligned}
\end{equation}
where $MHCA(\cdot)$ denotes the multi-head cross-attention layer, and $F_{c}^{\prime}$ is the intermediate features.
The evolved multi-modal feature $F_{c}$ is utilized for the final segmentation mask.
Note that the hyper-parameter $n$ is discussed in the following experiment section.

\subsection{Text-to-Pixel Contrastive Learning}
Although the CLIP \cite{radford2021learning} learns powerful image-level visual concepts by aligning the textual representation with the image-level representation,
this type of knowledge is sub-optimal for referring image segmentation, due to the lack of more fine-grained visual concepts.

To tackle this issue, we design a text-to-pixel contrastive loss,
which explicitly aligns the textual features with the corresponding pixel-level visual features.
As illustrated in Figure~\ref{fig:method}, image and text projector are adopted to transform $F_{c}$ and $F_{s}$ as follow:
\begin{equation}
\begin{aligned}
    &z_{v} = F_{c}^{\prime}W_{v}+b_{v},\,F_{c}^{\prime} = Up(F_{c}),\\
    &z_{t} = F_{s}W_{t}+b_{t},
    \label{eq8}
\end{aligned}
\end{equation}
where $z_{t} \in \mathbb{R}^{D}$, $z_{v} \in \mathbb{R}^{N \times D},\,N=\frac{H}{4} \times \frac{W}{4}$,
$Up$ denotes $4\times$ upsampling,
$W_{v}$ and $W_{t}$ are two learnable matrices to transform $F_{c}$ and $F_{s}$ into the same feature dimension $D$,
$b_{v}$ and $b_{t}$ are two learnable biases.

Given a transformed textual feature $z_{t}$ and a set of transformed pixel-level features $z_{v}$,
a contrastive loss is adopted to optimize the relationship between two modalities,
where $z_{t}$ is encouraged to be similar with its corresponding $z_{v}$ and dissimilar with other irrelevant $z_{v}$.
With the similarity measured by dot product, the text-to-pixel contrastive loss can be formulated as:
\begin{equation}
    L_{con}^{i}(z_{t},\,z_{v}^{i})=
    \begin{cases}
        -\log \sigma (z_{t} \cdot z_{v}^{i}), & i \in \mathcal{P},\\
        -\log (1 - \sigma (z_{t} \cdot z_{v}^{i})), & i \in \mathcal{N},\\
    \end{cases}
\end{equation}
\vspace{-4.0mm}
\begin{equation}
    L_{con}(z_{t},\,z_{v}) = \frac{1}{\left\lvert \mathcal{P} \cup \mathcal{N} \right\rvert} \sum_{i \in \mathcal{P} \cup \mathcal{N}} L_{con}^{i}(z_{t},\,z_{v}^{i}),
    \label{eq10}
\end{equation}
where $\mathcal{P}$ and $\mathcal{N}$ denote the class of ``1'' and ``0'' in the ground truth, $\left\lvert \mathcal{P} \cup \mathcal{N} \right\rvert$ is the cardinality, $\sigma$ is the sigmoid function.
Finally, to obtain the final segmentation results, we reshape $\sigma(z_{t} \cdot z_{v})$ into $\frac{H}{4} \times \frac{W}{4}$ and upsample it back to the original image size.


\begin{table*}[thbp]
    \setlength{\belowcaptionskip}{1.0pt}
    \begin{center}
    \caption{\textbf{Ablation studies on validation set of three benchmarks.}
    $Con.$ denotes the proposed text-to-pixel contrastive learning.
    $Dec.$ denotes the proposed vision-language decoder.
    $n$ denotes the number of layers in the vision-language decoder.
    We set $Num = 3$ as the default.
    ``Params'' and ``FPS'' denote the parameter complexity (M) and inference speed, respectively.
    Given an image $I \in \mathbb{R}^{416 \times 416 \times 3}$, they are calculated on a Tesla V100 GPU.
    \textcolor{gray}{Gray lines} denote the baseline network.}
    \setlength{\tabcolsep}{2mm}{
    \begin{tabular}{c|ccc|c|ccccc|cc}
        \toprule[1.2pt]
        Dataset & $Con.$ & $Dec.$ & $n$ & IoU & Pr@$50$ & Pr@$60$ & Pr@$70$ & Pr@$80$ & Pr@$90$ & Params & FPS \\
        \midrule[1.2pt]
        \multirow{4}{*}{RefCOCO}    & \textcolor{gray}{-} & \textcolor{gray}{-} & \textcolor{gray}{-} & \textcolor{gray}{62.66} & \textcolor{gray}{72.55} & \textcolor{gray}{67.29} & \textcolor{gray}{59.53} & \textcolor{gray}{43.52} & \textcolor{gray}{12.72} & \textcolor{gray}{131.86} & \textcolor{gray}{27.30} \\
         ~                          & \checkmark & - & - & 64.64 & 74.89 & 69.58 & 61.70 & 45.50 & 13.31 & 134.22 & 25.79 \\
         ~                          & - & \checkmark & 1 & 66.31 & 77.66 & 72.99 & 65.67 & 48.43 & 14.81 & 136.07 & 23.02 \\
         ~                          & \checkmark & \checkmark & 1 & 68.66 & 80.16 & 75.72 & 68.82 & 51.98 & 15.94 & 138.43 & 22.64 \\
         ~                          & \checkmark & \checkmark & 2 & 69.13 & 80.96 & 76.60 & 69.67 & 52.23 & 16.09 & 142.64 & 20.68 \\
         ~                          & \checkmark & \checkmark & 3 & \textbf{69.52} & \textbf{81.35} & \textbf{77.54} & \textbf{70.79} & \textbf{52.65} & 16.21 & 146.85 & 19.22 \\
         ~                          & \checkmark & \checkmark & 4 & 69.18 & 80.99 & 76.74 & 69.32 & 52.57 & \textbf{16.37} & 151.06 & 18.26 \\
        \midrule
        \multirow{4}{*}{RefCOCO+}   & \textcolor{gray}{-} & \textcolor{gray}{-} & \textcolor{gray}{-} & \textcolor{gray}{50.17} & \textcolor{gray}{54.55} & \textcolor{gray}{47.69} & \textcolor{gray}{40.19} & \textcolor{gray}{28.75} & \textcolor{gray}{8.21} & \textcolor{gray}{131.86} & \textcolor{gray}{27.30} \\
        ~                           & \checkmark & - & - & 53.15 & 58.28 & 53.74 & 46.67 & 34.01 & 9.30 & 134.22 & 25.79 \\
        ~                           & - & \checkmark & 1 & 54.73 & 63.31 & 58.89 & 52.46 & 38.53 & 11.70 & 136.07 & 23.02 \\
        ~                           & \checkmark & \checkmark & 1 & 59.97 & 69.19 & 64.85 & 58.17 & 43.47 & 13.39 & 138.43 & 22.64 \\
        ~                           & \checkmark & \checkmark & 2 & 60.75 & 70.69 & 66.83 & 60.74 & 45.69 & 13.42 & 142.64 & 20.68 \\
        ~                           & \checkmark & \checkmark & 3 & \textbf{61.39} & \textbf{71.46} & \textbf{67.82} & \textbf{61.80} & \textbf{47.00} & \textbf{15.02} & 146.85 & 19.22 \\
         ~                          & \checkmark & \checkmark & 4 & 61.15 & 71.05 & 66.94 & 61.25 & 46.98 & 14.97 & 151.06 & 18.26 \\
        \midrule
        \multirow{4}{*}{G-Ref}      & \textcolor{gray}{-} & \textcolor{gray}{-} & \textcolor{gray}{-} & \textcolor{gray}{49.24} & \textcolor{gray}{53.33} & \textcolor{gray}{45.49} & \textcolor{gray}{36.58} & \textcolor{gray}{23.90} & \textcolor{gray}{6.92} & \textcolor{gray}{131.86} & \textcolor{gray}{25.72} \\
        ~                           & \checkmark & - & - & 52.67 & 59.27 & 52.45 & 44.12 & 29.53 & 8.80 & 134.22 & 25.33 \\
        ~                           & - & \checkmark & 1 & 51.46 & 58.68 & 53.33 & 45.61 & 31.78 & 10.23 & 136.07 & 22.57 \\
        ~                           & \checkmark & \checkmark & 1 & 57.82 & 66.28 & 60.99 & 53.21 & 38.58 & 13.38 & 138.43 & 22.34 \\
        ~                           & \checkmark & \checkmark & 2 & 58.40 & 67.30 & 61.72 & 54.70 & 39.67 & 13.40 & 142.64 & 20.61 \\
        ~                           & \checkmark & \checkmark & 3 & \textbf{59.35} & \textbf{68.93} & \textbf{63.66} & \textbf{55.45} & \textbf{40.67} & \textbf{14.40} & 146.85 & 19.14 \\
         ~                          & \checkmark & \checkmark & 4 & 58.79 & 67.91 & 63.11 & 55.43 & 39.81 & 13.48 & 151.06 & 17.84\\
        \bottomrule[1.2pt]
    \end{tabular}
    \label{tab:ablation}}
    \end{center}
    \vspace{-5.0mm}
\end{table*}

\section{Experimental results}\label{sec:Experiments}
Our proposed framework is built on different image encoders (\eg, ResNet-50, ResNet-101 \cite{he2016deep}) and compared with a series of state-of-the-art methods.
To evaluate the effectiveness of each component in our method, we conduct extensive experiments on three benchmarks, including RefCOCO \cite{kazemzadeh2014referitgame}, RefCOCO+ \cite{kazemzadeh2014referitgame}, and G-Ref \cite{mao2016generation}.

\subsection{Datasets}
\textbf{RefCOCO} \cite{kazemzadeh2014referitgame} is one of the largest and most commonly used datasets for referring image segmentation.
It contains 19,994 images with 142,210 referring expressions for 50,000 objects,
which are collected from the MSCOCO \cite{lin2014microsoft} via a two-player game \cite{kazemzadeh2014referitgame}.
The dataset is split into 120,624 train, 10,834 validation, 5,657 test A, and 5,095 test B samples, respectively.
According to statistics, each image contains two or more objects and each expression has an average length of 3.6 words.

\textbf{RefCOCO+} \cite{kazemzadeh2014referitgame} dataset contains 141,564 language expressions with 49,856 objects in 19,992 images.
The dataset is split into train, validation, test A, and test B with 120,624, 10,758, 5,726, and 4,889 samples, respectively.
Compared with RefCOCO dataset, some kinds of absolute-location words are excluded from the RefCOCO+ dataset, which could be more challenging than the RefCOCO dataset.

\textbf{G-Ref} \cite{nagaraja2016modeling} includes 104,560 referring expressions for 54,822 objects in 26,711 images.
Unlike the above two datasets, natural expressions in G-Ref are collected from Amazon Mechanical Turk instead of a two-player game.
The average length of sentences is 8.4 words, which have more words about locations and appearances.
It is worth mentioning that we adopt UNC partition in this paper.

\subsection{Implementation Details}
\textbf{Experimental Settings.} 
We initiate the text and image encoder with CLIP \cite{radford2021learning}, and adopt ResNet-50 \cite{he2016deep} as the image encoder for all ablation studies.
Input images are resized to $416 \times 416$.
Due to the extra \texttt{[SOS]} and \texttt{[EOS]} tokens, and the input sentences are set with a maximum sentence length of 17 for RefCOCO and RefCOCO+, and 22 for G-Ref.
Each Transformer Decoder layer has 8 heads, and the feed-forward hidden dimension is set to 2048.
We train the network for 50 epochs using the Adam optimizer with the learning rate $\lambda = 0.0001$.
The learning rate is decreased by a factor of 0.1 at the 35th epoch.
We train the model with a batch size of 64 on 8 Tesla V100 with 16 GPU VRAM.

During inference, we upsample the predicted results back to the original image size and binarize them at a threshold of 0.35 as the final result.
No other post-processing operations are needed.

\textbf{Metrics.}
Following previous works \cite{ding2021vlt,liu2017recurrent,li2018referring,margffoy2018dynamic},
we adopt two metrics to verify the effectiveness: IoU and Precision@$X$.
The IoU calculates intersection regions over union regions of the predicted segmentation mask and the ground truth.
The Precision@$X$ measures the percentage of test images with an IoU score higher than the threshold $X\in\{0.5, 0.6, 0.7, 0.8, 0.9\}$, which focuses on the location ability of the method.

\subsection{Ablation Study}
The proposed CRIS framework consists of two main parts, \ie, text-to-pixel contrastive learning and vision-language decoder.
To investigate each component in our method, we conduct extensive experiments on the validation set of three widely used datasets.

\begin{table*}[thbp]
    \setlength{\belowcaptionskip}{1.0pt}
    \begin{center}
    \caption{\textbf{Comparisons with the state-of-the-art approaches on three benchmarks.}
    We report the results of our method with various visual backbones.
    ``$\star$'' denotes the post-processing of DenseCRF \cite{krahenbuhl2011efficient}.
    ``-'' represents that the result is not provided.
    IoU is utilized as the metric.}
    \setlength{\tabcolsep}{2.8mm}{
    \begin{tabular}{l|c|ccc|ccc|cc}
        \toprule[1.2pt]
        \multirow{2}{*}{Method} & \multirow{2}{*}{Backbone} & \multicolumn{3}{c|}{RefCOCO} & \multicolumn{3}{c|}{RefCOCO+} & \multicolumn{2}{c}{G-Ref} \\
        \cline{3-10}
        ~ & ~ & val & test A & test B & val & test A & test B & val & test \\
        \midrule[1.2pt]
        RMI$^\star$ \cite{liu2017recurrent}    & ResNet-101 & 45.18 & 45.69 & 45.57 & 29.86 & 30.48 & 29.50 & - & - \\
        DMN \cite{margffoy2018dynamic}         & ResNet-101 & 49.78 & 54.83 & 45.13 & 38.88 & 44.22 & 32.29 & - & - \\
        RRN$^\star$ \cite{li2018referring}     & ResNet-101 & 55.33 & 57.26 & 53.95 & 39.75 & 42.15 & 36.11 & - & - \\
        MAttNet \cite{yu2018mattnet}           & ResNet-101 & 56.51 & 62.37 & 51.70 & 46.67 & 52.39 & 40.08 & 47.64 & 48.61 \\
        NMTree \cite{liu2019learning}          & ResNet-101 & 56.59 & 63.02 & 52.06 & 47.40 & 53.01 & 41.56 & 46.59 & 47.88 \\
        CMSA$^\star$ \cite{ye2019cross}        & ResNet-101 & 58.32 & 60.61 & 55.09 & 43.76 & 47.60 & 37.89 & - & - \\
        Lang2Seg \cite{chen2019referring}      & ResNet-101 & 58.90 & 61.77 & 53.81 & - & - & - & 46.37 & 46.95 \\
        BCAN$^\star$ \cite{hu2020bi}           & ResNet-101 & 61.35 & 63.37 & 59.57 & 48.57 & 52.87 & 42.13 & - & - \\
        CMPC$^\star$ \cite{huang2020referring} & ResNet-101 & 61.36 & 64.53 & 59.64 & 49.56 & 53.44 & 43.23 & - & - \\
        LSCM$^\star$ \cite{huilinguistic}      & ResNet-101 & 61.47 & 64.99 & 59.55 & 49.34 & 53.12 & 43.50 & - & - \\
        MCN \cite{luo2020multi}                & DarkNet-53 & 62.44 & 64.20 & 59.71 & 50.62 & 54.99 & 44.69 & 49.22 & 49.40 \\
        CGAN \cite{luo2020cascade}             & DarkNet-53 & 64.86 & 68.04 & 62.07 & 51.03 & 55.51 & 44.06 & 51.01 & 51.69 \\
        EFNet \cite{feng2021encoder}           & ResNet-101 & 62.76 & 65.69 & 59.67 & 51.50 & 55.24 & 43.01 & - & - \\
        LTS \cite{jing2021locate}              & DarkNet-53 & 65.43 & 67.76 & 63.08 & 54.21 & 58.32 & 48.02 & 54.40 & 54.25 \\
        VLT \cite{ding2021vlt}                 & DarkNet-53 & 65.65 & 68.29 & 62.73 & 55.50 & 59.20 & 49.36 & 52.99 & 56.65 \\
        \midrule
        CRIS (Ours)                            & ResNet-50  & 69.52 & 72.72 & 64.70 & 61.39 & 67.10 & 52.48 & 59.35 & 59.39 \\
        CRIS (Ours)                            & ResNet-101 & \textbf{70.47} & \textbf{73.18} & \textbf{66.10} & \textbf{62.27} & \textbf{68.08} & \textbf{53.68} & \textbf{59.87} & \textbf{60.36} \\
        \bottomrule[1.2pt]
    \end{tabular}
    \label{tab:sota}}
    \end{center}
    \vspace{-5.0mm}
\end{table*}

\textbf{Effectiveness of Contrastive Learning \& Vision-Language Decoder.}
Firstly, we remove the parts of the text-to-pixel contrastive learning and vision-language decoder from the framework to build our baseline, which is same as the naive setting in Figure~\ref{fig:compare}.(c).
As illustrated in Table~\ref{tab:ablation}, we introduce the contrastive learning scheme, which significantly increases the IoU accuracy of 1.98\%, 2.98\%, and 3.43\% than the baseline network on three datasets, respectively.
This superior performance gain proves that the contrastive loss can encourage the model to explicitly pull closer linguistic and relevant pixel-level visual representations and push away other irrelevances for learning fine-structured multi-modal corresponding information.

Besides, we evaluate the performance of the proposed vision-language decoder. Compared with the baseline network, we use one layer in the decoder, bringing 3.65\%, 4.56\%, and 2.22\% IoU improvements on RefCOCO, RefCOCO+, and G-Ref, respectively.
In particular, the self-attention operation can help the model sufficiently capture long-range dependencies across each pixel, which is helpful for understanding complex scenarios.
Furthermore, each word encoded by the text encoder is used in the cross-attention operation, which can propagate fine-grained semantic information from textual features to pixel-level features to generate more discriminative visual representations and obtain more accurate segmentation masks.

Finally, combining the proposed contrastive loss and vision-language decoder, the IoU and Precision are significantly better than the baseline solely with the contrastive loss or decoder module, which further achieves large margins at about 4\% - 8\% on three datasets.
The reason of this obvious complementary phenomenon is that the contrastive loss can guide the decoder to find the more informative emphasis and transfer this knowledge to more accurate pixel-level visual representations, which boosts the ability of cross-modal matching and generates precise segmentation masks.

\textbf{Numbers of Layers in Decoder.}
In Table~\ref{tab:ablation}, the results illustrate the effect of utilizing different number of layers in the vision-language decoder.
When the visual representations are sequentially processed by more layers, our model can consistently get better IoU of 69.52\%, 61.39\%, and 59.35\% on three benchmarks, respectively.
The setting of $n=1$ may not taking full advantage of the multi-modal corresponding information from both vision and language.
Meanwhile, the setting of $n=4$ introduces more parameters, which could increase the risk of over-fitting.
Considering the performance and efficiency, we set $n=3$ as the default in our framework.

\subsection{Main Results}
We compare our proposed approach, CLIP-Driven Referring Image Segmentation, with a series of state-of-the-art methods on three commonly used datasets.
As illustrated in Table~\ref{tab:sota}, our method surpasses other methods on each split of all datasets even though we utilize a shallow ResNet-50 \cite{he2016deep}.

On the RefCOCO dataset, our model significantly outperforms the state-of-the-art Vision Language Transformer \cite{ding2021vlt} by 4.82\%, 4.89\% and 3.37\% on three splits, respectively, which indicates that our model effectively transfer the knowledge of the CLIP model from image-level to pixel-level, enhancing the ability of cross-modal matching.

Besides, in Table~\ref{tab:sota}, our method achieves remarkable performance gains of about 4\textasciitilde8\% than a series of state-of-the-art works on the more challenging RefCOCO+ dataset.
These obvious improvements over them suggest that our method can adequately leverage the powerful knowledge of the CLIP to accurately focus the region referred by the given language expression.

Furthermore, on another more complex G-Ref dataset where the average length of referring expressions is complicated, our proposed method consistently achieve notable improvement of around 5\% IoU than the state-of-the-art Locate then Segmentation \cite{jing2021locate}.
As shown in Table~\ref{tab:sota}, the results demonstrate that our proposed approach manages to understand long and complex sentences that contain more information and more emphases, and simultaneously perceive the corresponding object.
Apart from that, longer referring expressions could contain complex scenarios, which need a strong ability to model the global contextual information. Our proposed vision-language decoder is suitable to enhance the holistic understanding of vision and language features.

\begin{figure*}[t]
    \begin{center}
        \includegraphics[width=.92\linewidth]{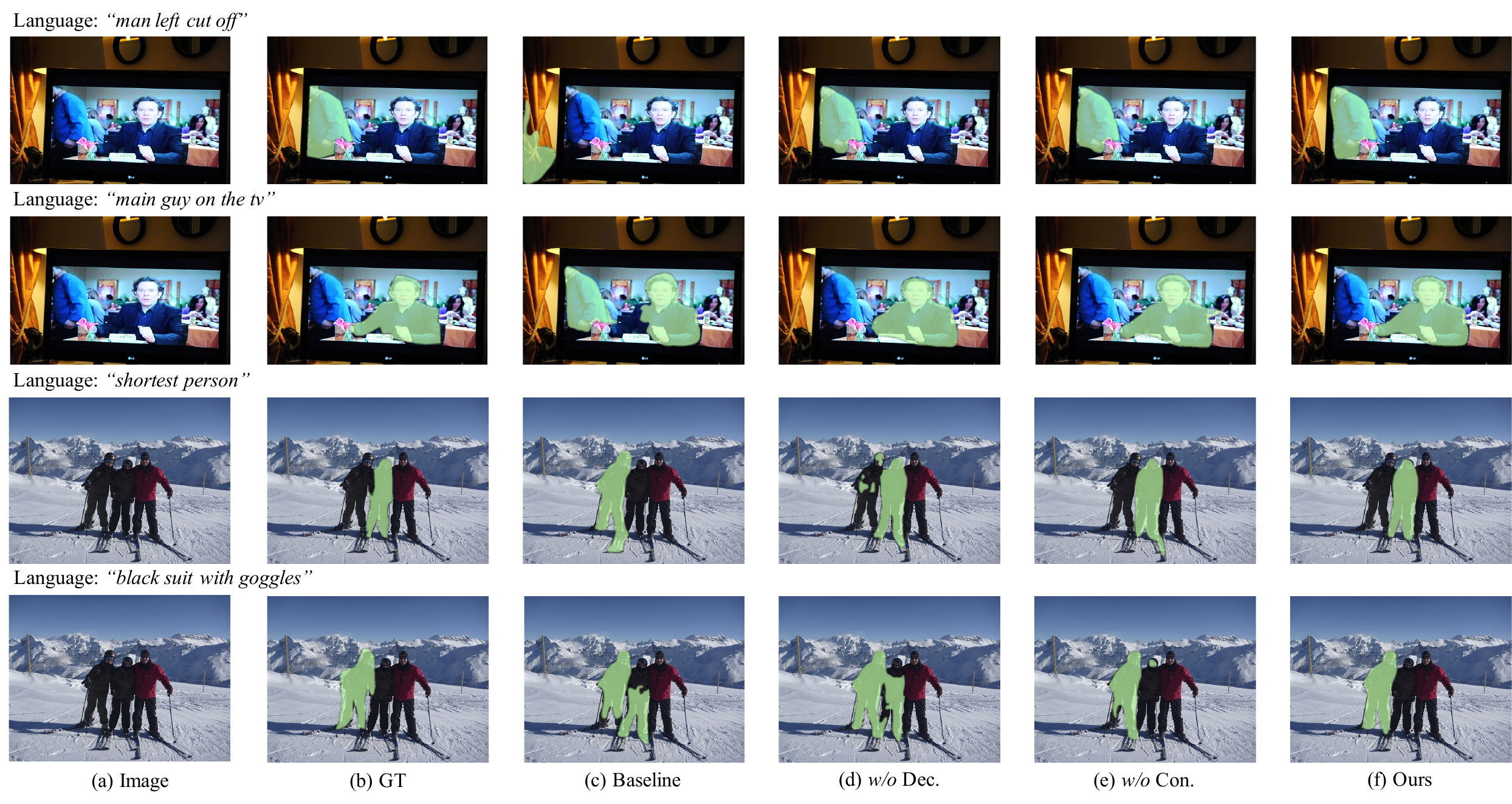}
        \caption{\textbf{Qualitative examples with different settings.}
        (a) the input image.
        (b) the ground turth.
        (c) the baseline network.
        (d) CRIS without Vision-Language Decoder.
        (e) CRIS without Contrastive Learning.
        (f) our proposed CRIS.
        \textit{Best viewed in color.}}
    \label{fig:visualize}
    \end{center}
    \vspace{-7.0 mm}
\end{figure*}
\begin{figure*}[t]
    \begin{center}
        \includegraphics[width=.921\linewidth]{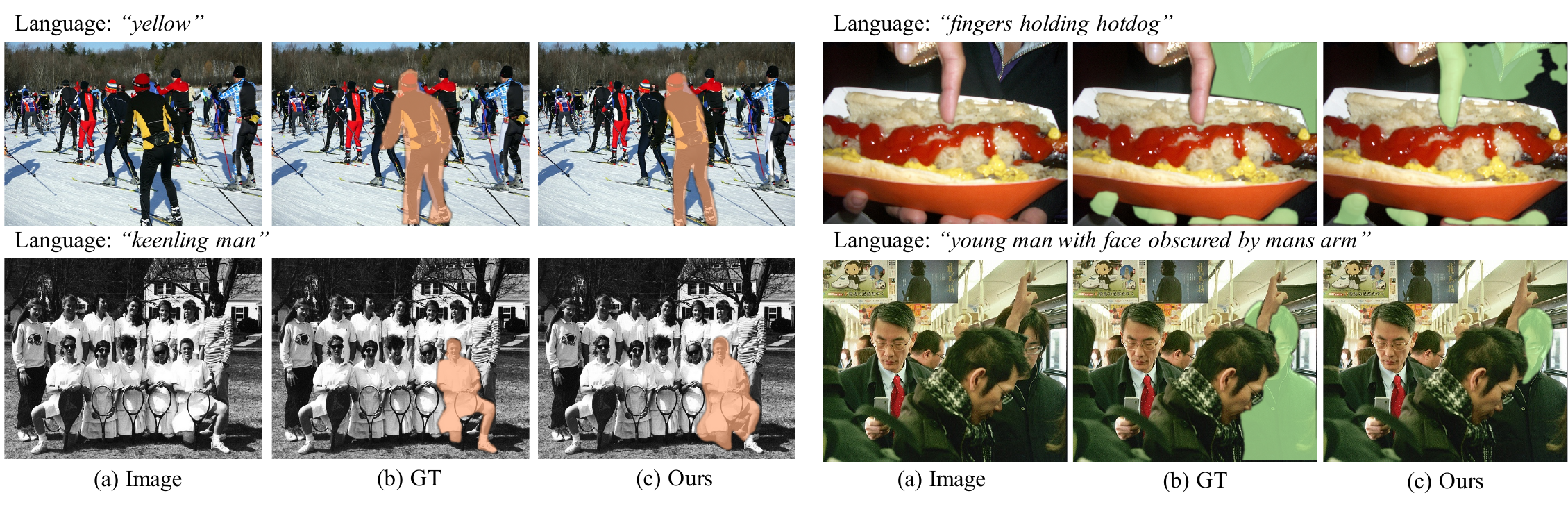}
        \caption{\textbf{Qualitative examples of failure cases.}
        \textit{Best viewed in color.}}
    \label{fig:bad_case}
    \end{center}
    \vspace{-7.0mm}
\end{figure*}

\subsection{Qualitative Analysis}
\textbf{Visualization.} As illustrated in Figure~\ref{fig:visualize}, we present some visualization results with different setting, which demonstrates the benefits of each component in our proposed method.
Firstly, compared with our full model, the baseline network without the contrastive learning and vision-language decoder generates worse segmentation masks, because the baseline network fails to interwind referring expressions with the corresponding regions.
Secondly, the setting of (d) and (e) can obtain more accurate segmentation results than the baseline network, but the model is still confused in some hard regions.
Finally, our proposed full model can generate high-quality segmentation masks, which demonstrates the effectiveness of our proposed method, \ie, CRIS.

\textbf{Failure Cases.} We visualize some insightful failed cases in Figure~\ref{fig:bad_case}.
One type of failure is caused by the ambiguity of the input expression. For the top left example in Figure~\ref{fig:bad_case}, the expression of ``yellow'' is not enough to describe the region of the man in the yellow snowsuit.
Besides, for the top right example, some failures are also caused by the wrong label. It is obvious that the top region is unrelated to ``fingers''.
As shown in the bottom left example, the boundaries of the referent cannot be accurately segmented, but this issue can be alleviated by introducing other technologies, such as the refine module.
Finally, occlusion could cause failure cases, which is a challenging problem in many vision tasks. 


\section{Conclusion}
In this paper, we have investigated to leverage the power of Contrastive Language-Image Pretraining (CLIP) models to achieve text-to-pixel alignment for referring image segmentation.
And, we have proposed an end-to-end CLIP-Driven Referring Image Segmentation (CRIS) framework to well transfer the knowledge of the CLIP model.
Compared with the direct fine-tuning, our proposed framework not only inherit the strong cross-modal matching ability of the CLIP, but also learn ample fine-structured visual representations.
The designed vision-language decoder can adaptively propagate sufficient semantic information of the language expression into pixel-level visual features, promoting consistency between two modalities.
Furthermore, the introduced text-to-pixel contrastive learning can explicitly interwind the text representation and relevant pixel-level visual features, learning fine-grained multi-modal corresponding information.
Extensive ablation studies on three commonly used datasets have verified the effectiveness of each proposed component,
and our approach significantly outperforms previous state-of-the-art methods without any post-processing.

\textbf{Acknowledgements}\quad We thank Ziyu Chen for the helpful discussions on this work, Weiqiong Chen, Bin Long and Rui Sun for AWS technical support.


{\small
  \bibliographystyle{ieee_fullname}
  \bibliography{mainbib}
}



\end{document}